\documentclass{article}
\usepackage{graphicx}
\usepackage[a4paper, left=1in, right=1in, top=1in, bottom=1in]{geometry}
\usepackage{tabularx}
\usepackage{amsmath, amssymb}
\usepackage{algorithm}
\usepackage{algpseudocode}
\usepackage{authblk}
\usepackage{float}
\usepackage[
    colorlinks=true,
    urlcolor=blue,
    citecolor=black,
    linkcolor=black,
    pdfborder={0 0 0}
]{hyperref}
\usepackage[
    backend=biber,
    style=numeric-comp,
    sorting=none,
    maxnames=2,
    minnames=1,
    giveninits=true
]{biblatex}
\usepackage{array}
\usepackage{caption}

\newcolumntype{P}[1]{>{\centering\arraybackslash}p{#1}}
\newcolumntype{C}{>{\centering\arraybackslash}X}

\DeclareNameAlias{default}{family-given}

\AtEveryBibitem{%
  \clearlist{editor}%
  \clearlist{publisher}%
  \clearfield{location}%
  \clearfield{isbn}%
  \clearfield{issn}%
  \clearfield{doi}%
  \clearfield{eprint}%
  \clearfield{booktitle}%
  \clearfield{eventtitle}%
  \clearfield{venue}%
}

\newcommand{\email}[2]{%
$^{#1}$\href{mailto:#2}{\texttt{\detokenize{#2}}}%
}

\title{Mitigating LLM Hallucinations through Domain-Grounded Tiered Retrieval}
\author[a,b,1]{Md. Asraful Haque}
\author[b,2]{Aasar Mehdi}
\author[b,3]{Maaz Mahboob}
\author[c,4]{Tamkeen Fatima}

\affil[a]{Computational Unit, Z.H. College of Engineering \& Technology, Aligarh Muslim University, India}
\affil[b]{Interdisciplinary Center for Artificial Intelligence, Aligarh Muslim University, India}
\affil[c]{Department of Computer Engineering, Aligarh Muslim University, India}

\date{Email: \email{1}{md_asraf@zhcet.ac.in}, \email{2}{asarmehdibaquri@gmail.com}, \email{3}{maazmahboob4@gmail.com}, \email{4}{tamkeen.fatima@zhcet.ac.in}}

\begin{document}

\maketitle
\begin{abstract}
\textbf{Background}: Large Language Models (LLMs) have achieved unprecedented fluency but remain susceptible to ``hallucinations"—the generation of factually incorrect or ungrounded content. This limitation is particularly critical in high-stakes domains where reliability is paramount.

\textbf{Objective}: We propose a domain-grounded tiered retrieval and verification architecture designed to systematically intercept factual inaccuracies by shifting LLMs from stochastic pattern-matchers to verified truth-seekers.

\textbf{Methodology}: The proposed framework utilizes a four-phase, self-regulating pipeline implemented via LangGraph: (I) Intrinsic Verification with Early-Exit logic to optimize compute, (II) Adaptive Search Routing utilizing a Domain Detector to target subject-specific archives, (III) Refined Context Filtering (RCF) to eliminate non-essential or distracting information, and (IV) Extrinsic Regeneration followed by atomic claim-level verification. The system was evaluated across 650 queries from five diverse benchmarks: TimeQA v2, FreshQA v2, HaluEval General, MMLU Global Facts, and TruthfulQA.

\textbf{Results}: Empirical results demonstrate that the pipeline consistently outperforms zero-shot baselines across all environments. Win rates peaked at 83.7\% in TimeQA v2 and 78.0\% in MMLU Global Facts, confirming high efficacy in domains requiring granular temporal and numerical precision. Groundedness scores remained robustly stable between 78.8\% and 86.4\% across factual-answer rows.

\textbf{Conclusion}: While the architecture provides a robust fail-safe for misinformation, a persistent failure mode of ``False-Premise Overclaiming" was identified. These findings provide a detailed empirical characterization of multi-stage RAG behavior and suggest that future work should prioritize pre-retrieval ``answerability" nodes to further bridge the reliability gap in conversational AI.

\end{abstract}

\section{Introduction}
\label{sec:intro}
\begin{figure}
    \centering
    \includegraphics[width=\linewidth]{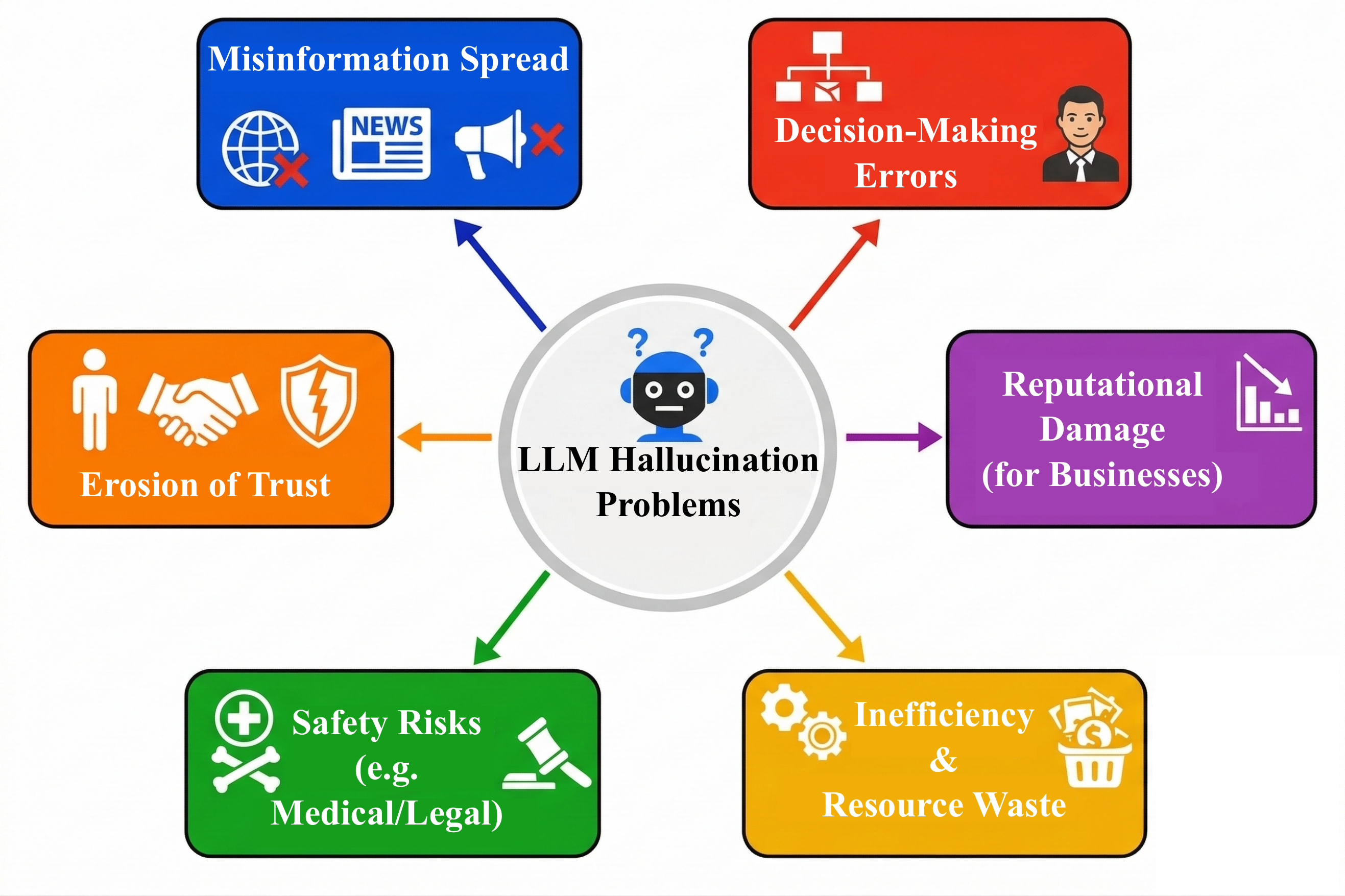}
    \caption{Consequences of LLM Hallucinations}
    \label{fig:consequences}
\end{figure}

Large Language Models (LLMs), including OpenAI’s ChatGPT, Google’s Gemini, and Meta’s Llama 3, have fundamentally transformed natural language processing. By training on massive datasets, these models have achieved a level of linguistic fluency often indistinguishable from human communication, excelling in tasks like reasoning, summarization, and creative writing \cite{r1,r2,r3}. Their rapid adoption across medicine, law, and journalism underscores their significant societal and industrial impact \cite{r2}. However, the credibility of LLMs is frequently undermined by hallucinations—a critical flaw where the model generates syntactically correct and coherent text that is despite being factually incorrect or unsupported by reliable sources \cite{r4,r5}. This issue is particularly severe in high-stakes environments where factual deviations can result in ethical breaches or dangerous decision-making errors \cite{r6} (Fig. \ref{fig:consequences}). The origins of hallucinations are multifactorial, stemming from data quality issues, the probabilistic nature of autoregressive generation, and ambiguous user prompts \cite{r7,r8}. Because LLMs are designed to predict the most likely next token rather than verify truth, they often prioritize plausibility over correctness \cite{r9}.

\begin{figure}[h]
    \centering
    \includegraphics[width=.65\linewidth]{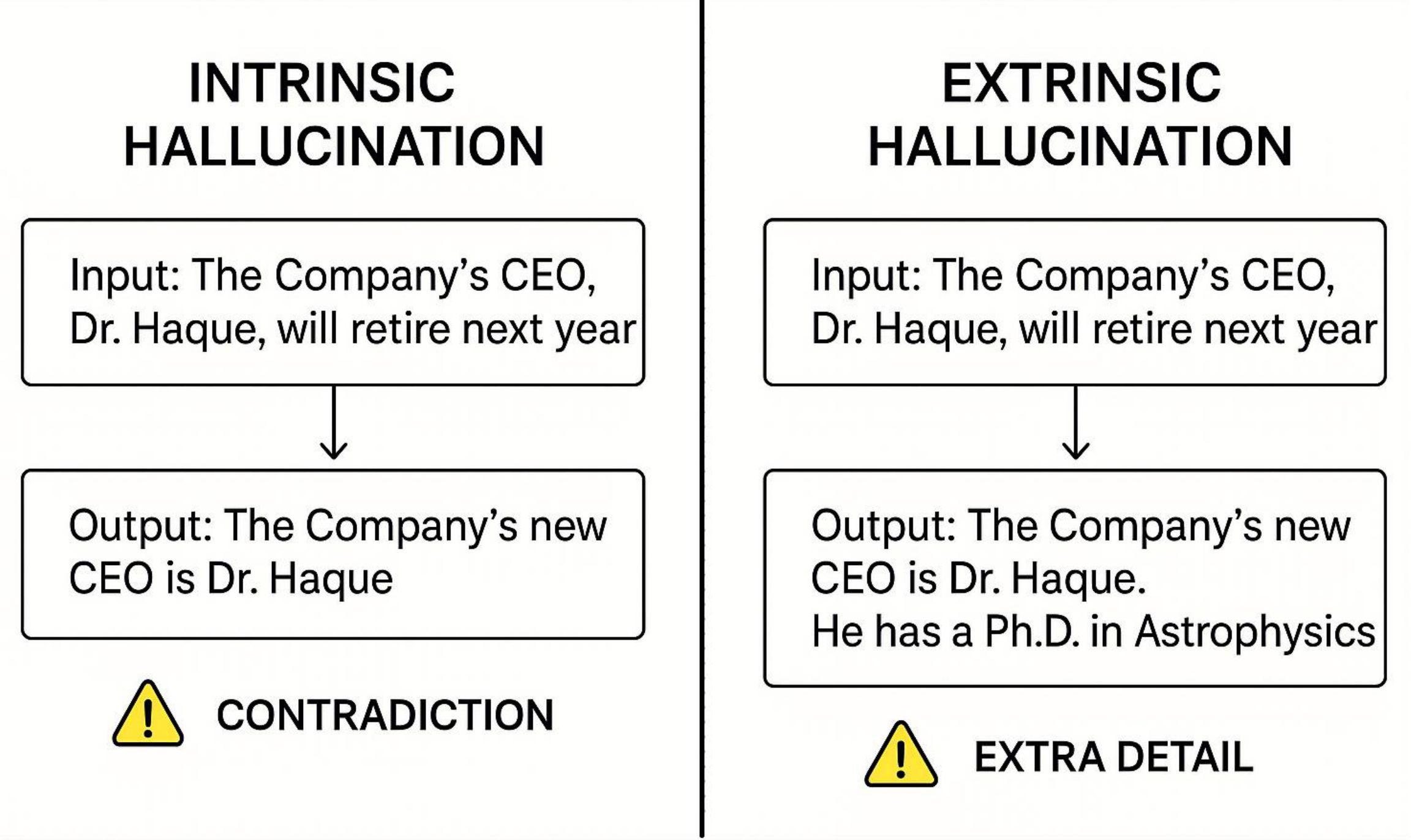}
    \caption{Intrinsic vs Extrinsic Hallucination}
    \label{fig:intrinsic_vs_extrinsic}
\end{figure}

Researchers generally categorize hallucination problems into two principal types \cite{r10}:
\begin{itemize}
    \item \textbf{Intrinsic Hallucinations}: These occur when a model’s internal limitations (such as data bias or restricted training diversity) cause it to distort or contradict the source information it is intended to process. This represents a failure of source fidelity, often resulting in outputs that directly oppose the provided input (Fig. \ref{fig:intrinsic_vs_extrinsic}).
    \item \textbf{Extrinsic Hallucinations}: These emerge when a model is queried on information absent from its training data or too recent for its internal knowledge base. Without access to external verification tools, the model relies on linguistic patterns to generate plausible but entirely unverifiable content, marking a fundamental failure in factual grounding (Fig. \ref{fig:intrinsic_vs_extrinsic}).
\end{itemize}

Beyond these two broad categories, hallucinations can also be sub-classified by the type of factual error they exhibit as shown in Table \ref{tab:hallu_types}. This fine-grained taxonomy helps identify the nature of the factual deviation, which is crucial for designing appropriate mitigation mechanisms.

\begin{table}[ht]
    \centering
    \caption{Hallucination Types based on Factual Deviation}
    \label{tab:hallu_types}
    \begin{tabularx}{\textwidth}{|p{2cm}|p{3cm}|X|p{3cm}|}
        \hline
        \multicolumn{1}{|c|}{\textbf{Type}} & \multicolumn{1}{|c|}{\textbf{Nature of Error}} & \multicolumn{1}{|c|}{\textbf{Example}} & \multicolumn{1}{|c|}{\textbf{Primary Cause}} \\
        \hline
        Factual & Incorrect statements contradicting truth & ``Albert Einstein won two Nobel Prizes in Physics." & Inaccurate memory / data bias \\
        \hline
        Numerical & Wrong figures or quantities & ``The Earth’s atmosphere contains 40\% oxygen." & Lack of numerical reasoning \\
        \hline
        Logical & Invalid reasoning chain & ``All doctors are scientists; Li is a scientist; therefore, Li is a doctor." & Weak inferential ability \\
        \hline
        Commonsense & Violates basic knowledge & ``The sun rises in the west and sets in the east." & Lack of world grounding \\
        \hline
        Temporal & Outdated or mixed-up timelines & ``Barack Obama is the current President of the United States." & Static knowledge \\
        \hline
        Attributional & Wrong author or source & ``The theory of relativity was proposed by Isaac Newton." & Entity confusion \\
        \hline
        Contextual & Factually true but irrelevant & In an article about climate change policy, the model inserts accurate facts about solar flares that are irrelevant to the policy discussion. & Poor discourse control \\
        \hline
        Referential & Fabricated entities or studies & ``According to a 2023 report by Wipro Tech., 500 million full-time jobs could be impacted by AI systems." (No such report exists) & Name pattern synthesis \\
        \hline
        Spatial & Wrong geographic facts & ``Sydney is capital of Australia." & Geographic misassociation \\
        \hline
        Multimodal & Cross-modal inconsistency & An image caption stating, ``A man playing guitar on stage," when no guitar is present. & Misaligned multimodal training \\
        \hline
    \end{tabularx}
\end{table}

AI shouldn't just sound smart; it also needs to be honest and clear about where its information comes from. When an AI model hallucinates, it ruins that trust. Recent benchmarks such as TruthfulQA show that even state-of-the-art models like GPT-3 achieve only $\sim$58\% truthfulness, while humans reach over 94\%, revealing a substantial reliability gap \cite{r11}. While strategies such as Retrieval-Augmented Generation (RAG) and Reinforcement Learning from Human Feedback (RLHF) have been introduced, significant limitations persist:
\begin{enumerate}
    \renewcommand{\labelenumi}{(\roman{enumi})}
    \item \textbf{Static Scrutiny}: Most frameworks apply the same level of verification to every response, leading to inefficiency and redundant checks for reliable claims.
    \item \textbf{Lack of Feedback Loops}: Current systems cannot dynamically adjust verification efforts based on internal confidence or past results \cite{r12,r13}.
    \item \textbf{Opacity}: Users are rarely informed about which parts of a response are verified or uncertain, reducing trust and interpretability.
    \item \textbf{Resource Intensity}: High-quality fact-checking often relies on expensive models (e.g., using GPT-4 to check GPT-3), making real-time deployment impractical \cite{r14,r15,r16}.
\end{enumerate}

To address these gaps, this paper introduces a multi-layered verification pipeline designed to be self-regulating, adaptive, and resource-efficient. This framework moves beyond standard ``search and answer" methods by incorporating internal and external checkpoints to ensure both fluency and factual integrity.

\section{Related Work}
\label{sec:rel_work}
Hallucination evaluation requires assessing whether generated statements are consistent with verifiable truth rather than simply fluent or coherent. Research on hallucinations in LLMs has grown rapidly, with efforts directed toward characterizing their occurrence, identifying their underlying mechanisms, and developing mitigation strategies. The first major approach to hallucination detection focuses on factual consistency—the degree to which the generated output aligns with source data or established truth. One of the earliest and most widely used tools is FactCC, a supervised model trained to detect factual inconsistencies in summarization outputs \cite{r17}. FactCC leverages a Natural Language Inference (NLI) framework to classify whether each generated sentence is supported, refuted, or unverifiable based on the source document. Despite its simplicity, FactCC’s binary nature limits its ability to handle nuanced or multi-sentence dependencies. To overcome these limitations, a later study introduced SummaC \cite{r18}. It utilizes a sentence-level entailment model that aggregates consistency judgments across all aligned text spans, thereby improving robustness in longer summaries and generative tasks. Several other widely used approaches for detecting hallucinations include truthfulness benchmarks, knowledge-grounding metrics, and Retrieval-Augmented Generation (RAG)–based methods. Ayala and Bechard \cite{r19} presented a RAG system that successfully reduced hallucinations in structured outputs, like workflows, that are generated from natural language. They developed a retriever that suggested steps from a knowledge base to enhance the language model's prompt and also enabled the system to generalize to new areas. However, Cheng et al. \cite{r20} showed that even with RAG, models can still generate unsupported information. They introduced RAGTruth, a new dataset of nearly 18,000 manually annotated responses to measure and address hallucinations in Retrieval-Augmented Generation (RAG) models. Another prevalent strategy for reducing hallucinations is to use Large Language Models (LLMs) to verify the output of other LLMs. While these LLM evaluators are efficient for checking general coherence or relevance, relying on them to judge factual accuracy is risky. This is because a faulty LLM may reinforce another's errors, creating an ``echo chamber" of misinformation without a ground truth. Tang et al. \cite{r21} addressed the high computational cost of fact-checking LLM outputs using other LLM like GPT-4. To solve this, they created MiniCheck, a small fact-checking model that achieves GPT-4-level performance at a cost that is 400 times lower. They accomplished this by generating a new synthetic training dataset using GPT-4, which helped the model learn to identify realistic and challenging factual errors. Ouyang et al \cite{r22} also noted that simply scaling up language models doesn't guarantee they will follow a user's intent, as they can produce unhelpful, untruthful, or toxic content. They introduced InstructGPT, a family of models fine-tuned on human feedback to better align with user intent. They collected a dataset of human rankings of model outputs and trained a reward model (RM) to predict which output a human would prefer. However, they acknowledged that InstructGPT still made mistakes. Farquhar et al. \cite{r23} developed a new, unsupervised technique to spot a particular kind of factual error in LLMs called ``confabulations." These errors are arbitrary, incorrect answers that are highly sensitive to irrelevant details, like a random seed. The central idea of the suggested technique is semantic entropy, which is a measure of uncertainty calculated based on the meaning of the generated sentences, not just the individual words or tokens. This is achieved by generating several answers to a question and then grouping them based on whether they share the same semantic equivalence. Lavrinovics et al. \cite{r24} addressed the key flaw of hallucinations in Large Language Models (LLMs) and proposed Knowledge Graphs (KGs) as a structured external knowledge source to mitigate this issue.

Some researchers believe that the key to eliminating hallucinations is to simply make models more accurate, assuming that a perfectly accurate model would never invent information \cite{r25}. However, this claim overlooks a fundamental limitation: 100\% accuracy is an impossible goal \cite{r26,r27}. No matter how large or capable a model becomes, it will always encounter questions that are inherently unanswerable in the real world. So, does this mean hallucinations are an inevitable side effect of language models? The answer is ``NO". We can teach LLMs to refrain from answering or simply reply with ``I don't know" when they lack sufficient information. Kalai et al. \cite{r25} modified standard evaluation benchmarks to encourage and reward Large Language Models (LLMs) for expressing uncertainty—for instance, by giving partial credit for responding with ``I don't know" or choosing to abstain from answering. They suggested adding explicit confidence targets to the evaluation prompts, which would guide models to only respond when they are sufficiently confident, thereby discouraging them from overconfident bluffing and, in turn, reducing hallucinations.

\section{Proposed Method: Domain-Grounded Tiered Retrieval Technique}
\label{sec:proposal}
To mitigate the persistent issue of hallucinations in LLMs, we propose a domain-grounded, tiered retrieval and verification architecture designed to detect and intercept the generation of factual inaccuracies. Instead of blindly searching the web for every question (which is expensive and slow), it acts like a human test-taker. First, it tries to answer from memory (closed-book). If it is absolutely certain it knows the answer, it stops. If it is uncertain or fails a strict constraint check, it opens the book (searches the web), but it does so in stages—checking trusted sources before falling back to the open web. If it still can't find the truth, it gracefully admits defeat rather than guessing. The proposed algorithm has been divided into four distinct phases as follows:

\begin{itemize}
    \item \textbf{Phase I: Intrinsic Verification and Early-Exit Logic} \\
        To optimize computational efficiency and reduce unnecessary API overhead, the system initially attempts to resolve the query using its internal parametric memory through a zero-shot baseline, $A_{init}$. This process begins by decomposing the generated response into discrete, verifiable atomic claims, which are then cross-referenced against the original query to identify potential constraint violations. If the response adheres to all predefined rules and constraints, a specialized Intrinsic Critic evaluates the factual reliability of the claims \cite{r28}. If the resulting internal confidence score $S_{intrinsic}$ meets or exceeds the established threshold $\tau$, the system triggers an early-exit mechanism, bypassing the retrieval pipeline to provide an immediate, high-confidence final response.
    \item \textbf{Phase II: Adaptive Search Routing and Domain-Specific Retrieval} \\
        When the initial intrinsic confidence score fails to meet the required threshold $\tau$, the system transitions from internal memory to an adaptive, tiered retrieval process designed to ground the response in external evidence. This phase begins with the Domain Detector, which classifies the user query $Q$ to identify the most relevant specialized archives, such as medical or legal databases. The routing logic then follows a prioritized hierarchy of trust, initially executing a TrustedSearch within these curated, subject-specific authorities to ensure maximum data integrity. If this primary tier fails to yield sufficient information, the system dynamically adapts by falling back to a GeneralWebSearch. This systematic progression through layered databases ensures that the model balances the precision of authoritative ``Gold Standard" sources with the broad coverage of the open web, only moving to less restricted tiers when specialized resources are exhausted.
    \item \textbf{Phase III: Refined Context Filtering (RCF)} \\
        Once the raw context $D_{raw}$ is retrieved from either the trusted or general search tiers, the system implements a Corrective Document Grading mechanism \cite{r29} to ensure the relevance and reliability of the external data. During this phase, a specialized grader evaluates each retrieved document against the original user query $Q$, filtering out noise, irrelevant snippets, or contradictory information to produce a refined set of high-quality context, $D_{filtered}$. If the grading process determines that the retrieved documents are empty or do not meet the necessary relevance criteria, the system triggers a recursive loop to the next search tier. This selective filtering prevents ``distractor" information from entering the generation stage, thereby significantly reducing the likelihood of retrieval-based hallucinations.
    \item \textbf{Phase IV: Extrinsic Regeneration and Verification} \\
        In the final stage of the pipeline, the system utilizes the refined context $D_{filtered}$ to perform Extrinsic Regeneration, where a new, grounded answer $A_{regen}$ is synthesized. To ensure absolute factual integrity, this regenerated response is decomposed into Atomic Claims \cite{r30} $C_{regen}$ which are then rigorously cross-referenced against the retrieved evidence. A secondary validation score $S_{retrieved}$ is calculated to measure how well the external data supports each individual claim. If this score meets the required confidence threshold $\tau$, the system outputs the verified RAG answer. However, if the verification fails or if all search tiers have been exhausted without finding corroborating evidence, a ``circuit breaker" is triggered to return a Graceful Apology, prioritizing the prevention of hallucination over the delivery of potentially false information.
\end{itemize}

\begin{algorithm}[H]
    \caption*{\textbf{Algorithm}}
    \begin{algorithmic}[1]
        \Statex \hspace*{-\algorithmicindent}\textit{\textbf{Input}: User Query Q, Confidence Threshold $\tau$ (e.g., 70)}
        \Statex \hspace*{-\algorithmicindent}\textit{\textbf{Output}: Final Factual Answer A*}
        \Statex \hspace*{-\algorithmicindent}\textit{\textbf{Initialize}: trusted\_done = False, general\_done = False}
        \Statex \begin{center}
                 \textbf{\textit{/* 1. Intrinsic Verification and Early Exit */}} 
            \end{center}
    
        \State $A_{init} \gets Generate(Q)$   \qquad \textit{// Zero-shot baseline}
        \State $C \gets ExtractClaims(A_{init})$
        \State $V \gets CheckConstraints(C,Q)$

        \If {\textit{V contains no constraint violations}}
            \State $S_{intrinsic} \gets ScoreIntrinsic(C)$   \qquad  \textit{// Closed-book evaluation}
            \If {$S_{intrinsic} \geq \tau$}
                \State \Return $A_{init}$   \qquad  \textit{// Bypass retrieval, exit early}
            \EndIf
        \EndIf

        \State $Domain \gets DetectDomain(Q)$
        
        \While{$True$}
            \Statex \begin{center}
                     \textbf{\textit{/* 2. Adaptive Search Routing */}} 
                \end{center}
            \If {\textit{not trusted\_done} }
                \State $D_{raw} \gets TrustedSearch(Q, Domain)$
                \State $trusted\_done \gets True$
            \ElsIf {\textit{not general\_done }}
                \State $D_{raw} \gets GeneralWebSearch(Q)$
                \State $general\_done \gets True$
            \Else
                \State \Return \textit{GenerateApology()}    \qquad \textit{// Circuit breaker triggered}
            \EndIf
            
            \Statex \begin{center}
                     \textbf{\textit{/* 3. Refined Context Filtering */}} 
                \end{center}
            \State $D_{filtered } \gets GradeDocuments(D_{raw}, Q)$
            \If {$D_{filtered }$ \textit{is empty}}
                \State \textit{continue}   \qquad  \textit{// Loop to next search tier}
            \EndIf
            
            \Statex \begin{center}
                     \textbf{\textit{/* 4. Extrinsic Regeneration \& Verification */}} 
                \end{center}
            \State $A_{regen} \gets Regenerate(Q, D_{filtered})$
            \State $C_{regen} \gets ExtractClaims(A_{regen})$
            \State $S_{retrieved } \gets ScoreRetrieved(C_{regen}, D_{filtered})$
            \If {$S_{retrieved} \geq \tau$}
                \State \Return $A_{regen}$   \qquad  \textit{// Verified RAG answer}
            \ElsIf {\textit{trusted\_done and general\_done}}
                \State \Return \textit{GenerateApology()}    \qquad	\textit{// Exhausted all resources}
            \EndIf
        \EndWhile
    \end{algorithmic}
\end{algorithm}

\begin{figure}
    \centering
    \includegraphics[width=\linewidth]{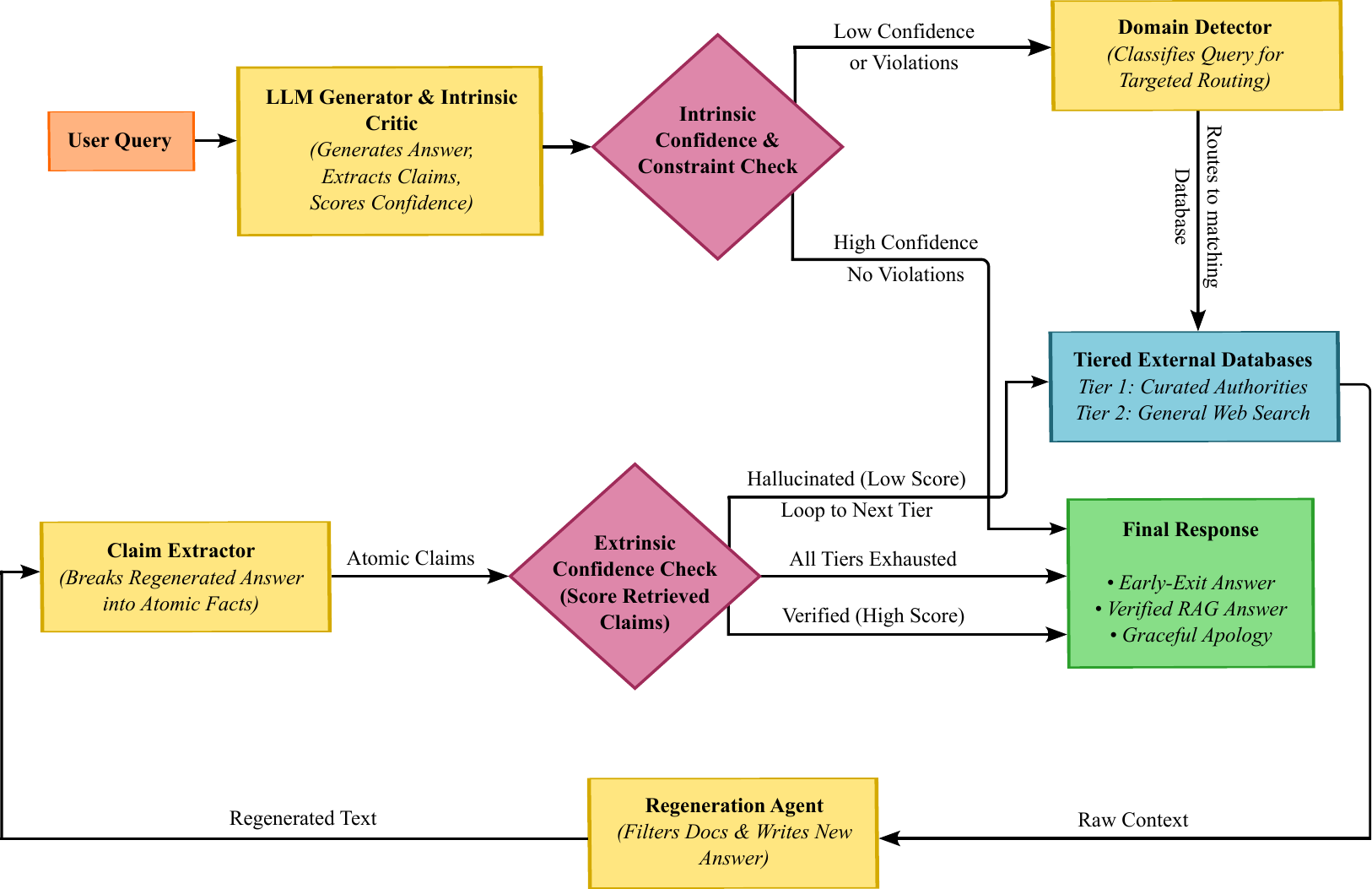}
    \caption{Proposed Framework}
    \label{fig:proposed_framework}
\end{figure}

Fig. \ref{fig:proposed_framework} depicts the workflow of the proposed method, beginning with an `Intrinsic Confidence \& Constraint Check' to assess if the model can generate an accurate response from its internal memory. If the confidence is low, a ``Domain Detector" classifies the query and routes it through ``Tiered External Databases", starting with curated authorities before falling back to a general web search. The architecture emphasizes verification by breaking regenerated answers into ``Atomic Claims" that undergo an ``Extrinsic Confidence Check" to score retrieved facts, ultimately leading to a verified RAG answer or a safe refusal if all tiers are exhausted.

\section{Performance Evaluation}
\label{sec:evaluation}
This section details the experimental configuration, the benchmarks utilized for stress-testing the architecture, and the metrics employed to quantify hallucination mitigation.

\subsection{Experimental Configuration}
\label{subsec:config}
The evaluation framework is constructed as a multi-stage directed graph implemented in LangGraph. Each node corresponds to a distinct processing step, with conditional routing edges determining whether the pipeline escalates from parametric memory to retrieval-grounded generation. Most internal tasks, including answer generation, atomic claim extraction, and document relevance grading, are executed by Llama 3.1 8B. This model is deployed locally via Ollama with a temperature of 0 to ensure deterministic and consistent outputs. To keep the data organized and accurate during the checking phase, the system uses Pydantic for structured validation. External data retrieval is facilitated via the Tavily search API, which queries both subject-specific trusted repositories and the broader web. For the final evaluation, Gemma3 27B serves as an independent, asymmetric judge to compare the pipeline's performance against a zero-shot baseline.

\subsection{Benchmark Selection and Taxonomy}
\label{subsec:benchmarks}
The proposed pipeline was evaluated across 650 queries derived from five diverse benchmarks, specifically selected to stress-test the architecture's ability to handle distinct factual deviations, temporal constraints, and logical complexities (Table \ref{tab:benchmarks}). The selection includes TimeQA v2 \cite{r31}, which targets temporal biographical reasoning through questions regarding roles held within specific date ranges, often characterized by sparse documentation. FreshQA v2 \cite{r32} assesses the retrieval of current knowledge and robustness against false premises. To evaluate broader capabilities, HaluEval General \cite{r33} provides a large-scale assessment of open-domain hallucinations in science, coding, and general knowledge, while MMLU Global Facts \cite{r34} focuses on precise numerical statistics that typically exceed the parametric memory of smaller 8B models. Finally, TruthfulQA \cite{r11} is employed to measure the system’s resistance to popular myths and common misconceptions, ensuring the architecture can identify and deny false premises rather than succumbing to confabulation.

\begin{table}[ht]
    \caption{Benchmark summary}
    \label{tab:benchmarks}
    \begin{tabularx}{\textwidth}{|P{3cm}|P{.5cm}|P{2.5cm}|C|C|}
        \hline
        \textbf{Benchmark} & \textbf{N} & \textbf{Domain} & \textbf{Query Type} & \textbf{Key Challenges} \\
        \hline
        TimeQA v2 & 150 & Temporal biographical & `Who held role X between dates A–B?' & Sparse time-window documentation; correct refusal on unknowable queries \\
        \hline
        FreshQA v2 & 150 & Current events + false premises & Factual + denial queries & 39\% of queries embed false or unresolved premises; requires refusal, not inference \\
        \hline
        HaluEval General & 150 & Open domain (mixed) & General knowledge, science, coding & Broad coverage; retrieval advantage lower vs. parametrically-strong baseline \\
        \hline
        MMLU Global Facts & 50 & Global statistics & Precise numerical recall & Exact figures unavailable in small-model memory; short answers \\
        \hline
        TruthfulQA & 150 & Myths / misconceptions & Myth-busting; denial of false premises & Must resist confabulation; false-premise resistance essential \\
        \hline
    \end{tabularx}
\end{table}

\subsection{Evaluation Metrics}
\label{subsec:metrics}
We use several key metrics to determine how well the new RAG system works compared to a standard model. Below is a detailed explanation of these metrics:

\begin{enumerate}
    \renewcommand{\labelenumi}{(\roman{enumi})}
    \item \textbf{Primary Performance Metrics} \\
        \begin{itemize}
            \item \textbf{Win Rate}: This is the percentage of total queries where the independent judge (Gemma3 27B) preferred the answer generated by the RAG pipeline over the standard, ``zero-shot" answer provided by Llama 3.1 8B. A higher win rate proves that the retrieval steps successfully added valuable information that the model did not already know.
            \item \textbf{Groundedness Score}: This metric tracks the portion of individual factual statements (atomic claims) in the final answer that are directly supported by the retrieved documents. It serves as a measure of how well the model actually uses the evidence it finds; therefore, a higher score is better.
            \item \textbf{Hallucination Rate}: This is the opposite of groundedness. It represents the fraction of factual claims made by the model that cannot be found in or verified by the retrieved evidence. To ensure the system is reliable and not ``making things up," a lower value is better.
        \end{itemize}
    \item \textbf{Supplementary Metrics} \\
        \begin{itemize}
            \item \textbf{Tie Rate}: This indicates the frequency with which the judge found both the pipeline's answer and the baseline answer to be of equal quality. This often happens when the model is already confident and skips retrieval, or when both models correctly identify that a question is unanswerable.
            \item \textbf{Baseline Win Rate}: This records the instances where the judge actually preferred the simple, zero-shot answer over the one provided by the complex RAG pipeline. This usually occurs due to ``overclaiming," where the pipeline tries to force a factual answer onto a trick question or a false premise.
        \end{itemize}
\end{enumerate}

\section{Discussion and Analysis}
\label{sec:discussion}
\begin{figure}
    \centering
    \includegraphics[width=\linewidth]{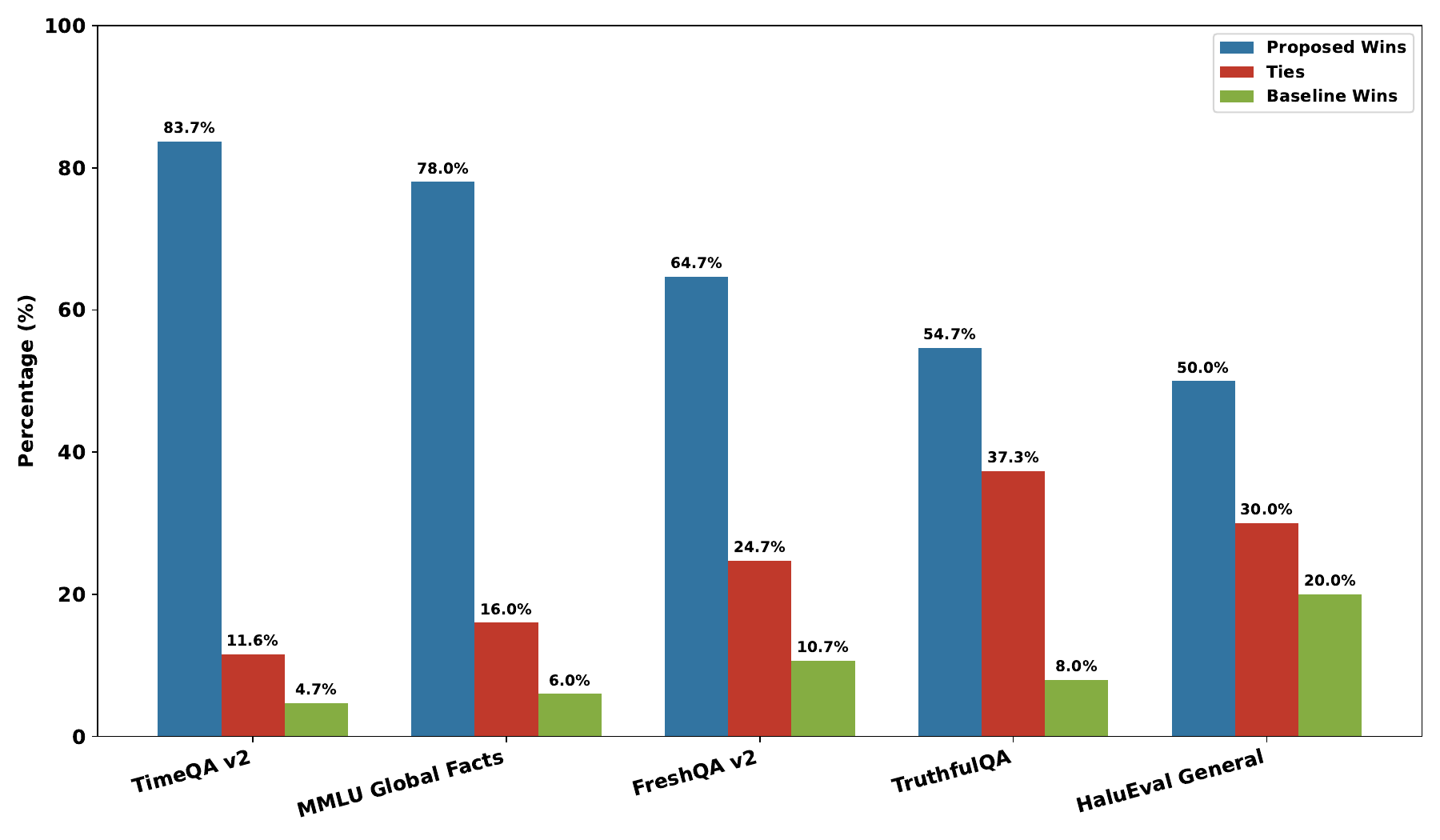}
    \caption{Benchmark Win Rate Comparison}
    \label{fig:comparison}
\end{figure}

The proposed architecture consistently outperformed the zero-shot baseline across all tested environments, with win rates ranging from 50.0\% to 83.7\% (Table \ref{tab:res}). Fig. \ref{fig:comparison} illustrates the comparative performance of the proposed RAG pipeline versus the zero-shot baseline. This section analyzes the performance drivers, systemic failure modes, and the stability of groundedness across diverse NLP domains.

\subsection{Retrieval Advantage and Query Specificity}
\label{subsec:retrieval_adv}
The evaluation demonstrates that retrieval advantage is strongly correlated with the degree to which a query exceeds the model's internal training data. This relationship follows the ``specificity-advantage curve," where the pipeline performs best on highly specific queries, particularly those involving narrow temporal or numerical information. The curve is characterized by two factors: specificity (narrowness of the requested fact), and recency (likelihood of post-training information change).

\begin{itemize}
    \item \textbf{High-Advantage Queries}: The highest win rates were observed in TimeQA v2 (83.7\%) and MMLU Global Facts (78.0\%). These benchmarks require granular temporal biographical data or precise numerical statistics—information that is either absent or poorly encoded in a small model like Llama 3.1 8B.
    \item \textbf{Competitive Parametric Memory}: In HaluEval General, the win rate dropped to 50.0\%. This suggests that for common-knowledge or procedural tasks (e.g., coding or general science), the baseline's parametric memory is a strong competitor, and retrieval provides a more modest benefit.
    \item \textbf{Adaptive Efficiency}: The pipeline successfully bypassed retrieval for 20\% of HaluEval queries via the intrinsic halting criterion, proving the architecture can avoid unnecessary latency for creative or procedural tasks without sacrificing quality.
\end{itemize}

\begin{table}[ht]
    \caption{Primary results across all five benchmarks}
    \label{tab:res}
    \begin{tabularx}{\textwidth}{|p{2.25cm}|P{.5cm}|C|P{.7cm}|C|C|P{2.25cm}|P{2.5cm}|}
        \hline
        \multicolumn{1}{|c|}{\textbf{Benchmark}} & \textbf{N} & \textbf{Proposed Wins} & \textbf{Tie} & \textbf{Baseline Wins} & \textbf{Win Rate} & \textbf{Hallucination} & \textbf{Groundedness} \\
        \hline
        TimeQA v2 & 86* & 72 & 10 & 4 & 83.7\% & 13.6\% & 86.4\% \\
        \hline
        MMLU Global Facts & 50 & 39 & 8 & 3 & 78.0\% & 33.1\% & 66.9\% \\
        \hline
        FreshQA v2 & 150 & 97 & 37 & 16 & 64.7\% & 3.5\%† & 19.2\%† \\
        \hline
        TruthfulQA & 150 & 82 & 56 & 12 & 54.7\% & 15.1\% & 84.9\% \\
        \hline
        HaluEval General & 150 & 75 & 45 & 30 & 50.0\% & 21.2\% & 78.8\% \\
        \hline
        Combined (650) & 586 & 365 & 220** & 65 & 62.3\% & $-$ & $-$ \\
        \hline
    \end{tabularx}
    \vspace{0.01em}
    \footnotesize{
    \textbf{\textit{* TimeQA: 150 questions attempted; 86 valid pipeline runs (64 queries invalidated due to Tavily API quota exhaustion, produced no pipeline output and recorded as Ties). \\
    † FreshQA: N=150 figures include 116 correct denial answers (0/0 scores by design). Considering factual-answer subset (N=34): hallucination 15.5\% and groundedness 84.5\%. \\
    ** Combined Ties = 220 = 156 valid ties (10+8+37+56+45) + 64 API-exhausted TimeQA queries.}}
    }
\end{table}

\subsection{Groundedness Stability Across Domains}
\label{subsec:groundedness_stability}
While hallucination rates appeared to vary, groundedness—the fraction of claims supported by retrieved evidence—showed remarkable consistency.

\begin{itemize}
    \item \textbf{Consistent Performance}: On factual-answer rows, groundedness remained stable between 78.8\% and 86.4\% across four benchmarks. MMLU Global Facts represents an outlier at 66.9\%; however, this is attributable to measurement sensitivity rather than genuine grounding failure. Because these answers typically consist of only 2–3 atomic claims, minor discrepancies in units or data vintage between the retrieved source and the generated output disproportionately inflate the hallucination score.
    \item \textbf{Measurement Sensitivity in MMLU}: The higher reported hallucination rate in MMLU Global Facts (33.1\%) is attributed to measurement sensitivity rather than factual failure. Because numerical answers are short (2–3 atomic claims), any slight difference in units or data vintage between the retrieved source and the generated answer triggers a high hallucination score.
\end{itemize}

\subsection{Failure Mode Characterization}
\label{subsec:failure_modes}
The empirical evaluation of the 650 queries identified six primary categories of failure that account for the 65 competitive losses recorded against the baseline. Table \ref{tab:failure_modes} presents these failure modes with their frequency, affected benchmarks, and root causes. As certain losses exhibit multiple overlapping failure patterns, the cumulative frequency in the table reaches 74. Understanding these patterns is essential for transitioning the architecture toward a more robust, ``denial-aware" state machine.

\begin{enumerate}
    \renewcommand{\labelenumi}{(\roman{enumi})}
    \item \textbf{Open-Domain Parametric Competition}: This failure mode occurs when the baseline model's internal training data is already sufficient to provide a high-quality answer without external assistance. For well-established scientific facts or common historical knowledge, the retrieval process adds computational overhead without providing a measurable improvement in accuracy. This was most prevalent in the HaluEval benchmark, where the baseline often matched the quality of retrieved web sources. To address such failures, the system should use a dynamic weighting mechanism that compares the Intrinsic Confidence Score ($S_{intrinsic}$) with the potential information gain from retrieval. If the model is highly confident about a common-knowledge query, the system should prioritize a zero-shot response to reduce latency and avoid retrieval-induced noise.
    \item \textbf{False-Premise Overclaiming}: The system occasionally prioritizes contextually related information over the recognition of impossible or unanswerable query premises. This leads to the generation of a confident but factually incorrect response that validates a false query. In FreshQA, when asked about Lionel Messi winning a second World Cup, the pipeline retrieved data about the 2022 win but failed to notice the count was incorrect, ``confirming" the second win. Similarly, in TimeQA, the system inferred roles from adjacent biographical data rather than admitting a lack of evidence for a specific time window. A dedicated Answerability Node should act as a gatekeeper that checks whether a query is based on a valid premise. It should utilize Pydantic-based structured outputs to verify the truthfulness of the query. If the premise is false or impossible, the system redirects the query to a refusal-generation path instead of producing an overclaim.
    \item \textbf{Vagueness (Verbose vs. Concise Denial)}: Vagueness occurs when the generator identifies that an event did not happen but produces a long, heavily hedged response. While these multi-paragraph answers are factually defensible, evaluation judges often prefer the baseline's direct and simple ``No". This behavior is largely a consequence of the smaller 8B parameter model's limited ability to synthesize concise refusals under complex retrieval constraints. Such verbosity can reduce the perceived utility of the system for end-users seeking clear information. Therefore, scaling the generation nodes to a mid-size model (such as Llama 3.3 70B) is recommended. Larger models typically exhibit better instruction-following capabilities, allowing them to produce concise and direct denials without excessive hedging.
    \item \textbf{Retrieval Distraction}: In this failure mode, the system retrieves a document that contains tangential or irrelevant context which misleads the generator. Even if the correct answer is present in the source, the model may focus on a side topic and produce a response that diverges from the user's core intent \cite{r35}. This highlights a vulnerability in the model's attention mechanism when dealing with long or multi-topic retrieved passages. It demonstrates that more information does not always result in a more accurate output if the model cannot distinguish signal from noise. Enhancing the Refined Context Filtering logic is necessary to improve context isolation. Future work should also explore more aggressive ``literalness" instructions for the generator to ensure it ignores tangential data in favor of the primary factual evidence.
    \item \textbf{Numerical Precision and Data Mismatch}: This category involves failures where the retrieved source and the final generated claim use different units, aggregation methods, or timeframes. In MMLU Global Facts, a source might provide data from one year while the query asks for another, or use different statistical denominators. Because the system relies on exact matching for verification, these minor discrepancies result in high hallucination scores even when the answer is nearly correct. Therefore, evaluation metrics should transition from rigid keyword comparisons to numerical proximity scoring. This would allow the system to better accommodate small discrepancies in data chronology or precision that do not constitute a genuine hallucination.
    \item \textbf{Structured Data Extraction Errors}: Failures in this mode result from the model's inability to correctly interpret or extract values from structured documents like markdown tables or lists. The Llama 3.1 8B model occasionally misreads row/column relationships, leading to the extraction of incorrect data points. This is an inherent limitation of smaller models, which lack the sophisticated reasoning required to parse dense technical documentation. These errors suggest that scaling to larger models is necessary for tasks involving complex document structures.
\end{enumerate}

\begin{table}[ht]
    \centering
    \caption{Summary of Failure Modes}
    \label{tab:failure_modes}
    \begin{tabularx}{\textwidth}{|p{2.25cm}|P{1.75cm}|P{3.75cm}|X|}
        \hline
        \multicolumn{1}{|c|}{\textbf{Failure Mode}} & \multicolumn{1}{|c|}{\textbf{Frequency}} & \multicolumn{1}{|c|}{\textbf{Primary Benchmarks}} & \multicolumn{1}{|c|}{\textbf{Primary Root Cause}} \\
        \hline
        Open-Domain Parametric Competition & $\sim$30 & HaluEval & High-quality training data in the baseline model matches the quality of retrieved web sources for common knowledge. \\
        \hline
        False-Premise Overclaiming & $\sim$14 & TimeQA, FreshQA, TruthfulQA & Lack of a pre-retrieval answerability check; the system answers the premise instead of correcting it. \\
        \hline
        Vagueness (Verbose vs. Concise) & $\sim$12 & FreshQA, TruthfulQA & The generator produces long, hedged responses that the judge finds less useful than the baseline's concise denial. \\
        \hline
        Retrieval Distraction & $\sim$9 & TruthfulQA, HaluEval & Tangential information in a retrieved document leads the generator away from the direct answer. \\
        \hline
        Numerical precision / Data mismatch & $\sim$5 & MMLU, FreshQA & The retrieved source utilizes different aggregation methods, timeframes, or units than the generated claim. \\
        \hline
        Structured Data Extraction Errors & $\sim$4 & FreshQA, MMLU & Small-model limitations (Llama 3.1 8B) leading to misreading of markdown tables or extracting wrong values from documents. \\
        \hline
    \end{tabularx}
\end{table}

\begin{figure}
    \centering
    \includegraphics[width=\linewidth]{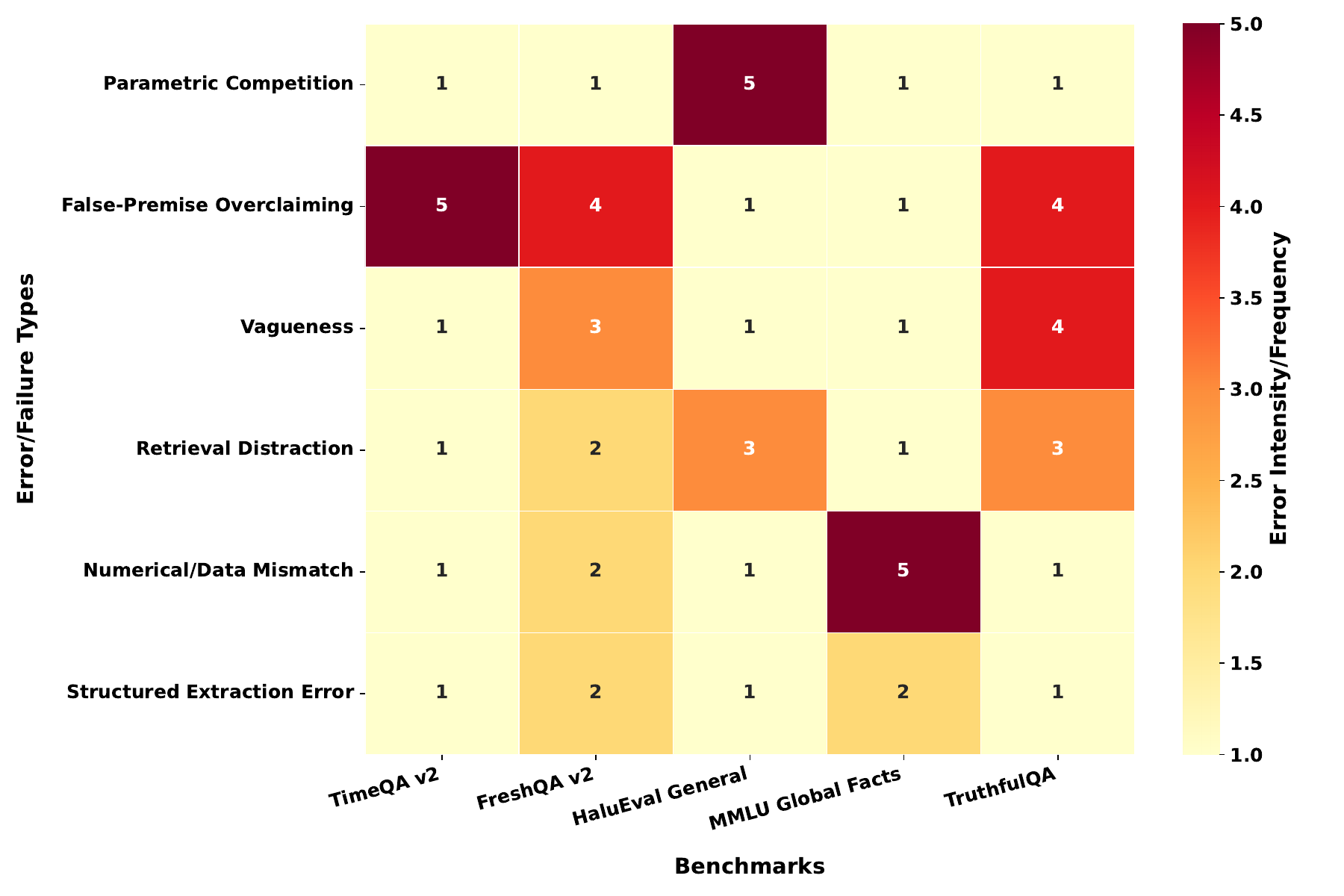}
    \caption{Realibility Heatmap (Benchmarks vs Error Types)}
    \label{fig:heatmap}
\end{figure}

Fig. \ref{fig:heatmap} provides provides a granular visualization of how different failure modes cluster across our chosen benchmarks. The heatmap reveals localized ``stress points" in the architecture, allowing for targeted optimization. False-Premise Overclaiming and Vagueness are most intense in refusal-heavy benchmarks like TruthfulQA and FreshQA v2. Conversely, Numerical/Data Mismatch is almost exclusively concentrated in MMLU Global Facts, highlighting a measurement sensitivity issue in statistical queries. Additionally, the high intensity of Parametric Competition in HaluEval General validates the efficiency of the Intrinsic Halting Criterion, showing where internal memory is already sufficient. Ultimately, the heatmap identifies specific targets for architectural remediation, such as the need for an ``Answerability Node" to handle the clusters of overclaiming errors.

\section{Conclusions}
\label{sec:conclusion}

The proposed algorithm functions as a robust, specialized retrieval framework designed for high-stakes accuracy. By grounding its outputs in verified, domain-specific repositories, the system delivers authoritative information instead of relying on broad, unverified web data. This tiered strategy prioritizes the most reliable sources first, incorporating a built-in safety protocol to provide a polite disclaimer if a factual basis cannot be confirmed. Evaluation of 650 queries across five diverse NLP benchmarks demonstrates that a multi-stage RAG pipeline significantly boosts the factual grounding of local models. The pipeline consistently and substantially outperforms the zero-shot baseline, particularly in domains demanding high specificity and temporal recency. Win rates peaked at 83.7\% for TimeQA v2 and 78.0\% for MMLU Global Facts, illustrating the pipeline's effectiveness in retrieving granular biographical and numerical data that typically exceeds a model's parametric memory. Conversely, the modest 50.0\% win rate in HaluEval General confirms that retrieval augmentation provides a more limited benefit for open-domain common knowledge already encoded in large-scale training data. A critical architectural success was the implementation of adaptive halting criteria. The intrinsic halting mechanism correctly bypassed retrieval for 20\% of open-domain queries involving procedural or creative tasks, thereby reducing latency and mitigating retrieval noise. Furthermore, the consistency of groundedness scores (78.8\% to 86.4\%) on factual queries across the benchmarks indicates that the pipeline's verification loops provide stable, domain-agnostic grounding quality. However, the evaluation also identified few failure patterns including the false-premise overclaiming issue. Across multiple benchmarks, the system occasionally prioritized contextually related information over recognizing impossible or unanswerable query premises. This highlights a specific need for a dedicated pre-retrieval answerability check in future iterations. Bridging the gap between human-level verification and automated generation is essential for neutralizing hallucinations. This shift will allow the AI community to create conversational models that are as credible and secure as they are articulate.

\printbibliography

\end{document}